\documentclass[a4paper,twocolumn,10pt]{article}
\usepackage{eurosis}
\usepackage{url}
\usepackage{natbib}
\usepackage{float}
\usepackage{graphicx}

\bibpunct[; ]{(}{)}{,}{a}{}{;}

\begin{document}

\title{\MakeUppercase{A Multi-Agent System for Autonomous Mobile Robot Coordination}}

\author{Norberto Sousa\textsuperscript{[0000-0003-2919-4817]}, Nuno Oliveira\textsuperscript{[0000-0002-5030-7751]} \\ and Isabel Praça\textsuperscript{[0000-0002-2519-9859]} \\
Research Group on Intelligent Engineering and Computing for Advanced Innovation and Development (GECAD),\\ Porto School of Engineering (ISEP), 4200-072 Porto, Portugal \\
\texttt{\{norbe,nunal,icp\}@isep.ipp.pt}}

\date{}

\maketitle

\thispagestyle{empty}


\begin{abstract}
The automation of internal logistics and inventory-related tasks is one of the main challenges of modern-day manufacturing corporations since it allows a more effective application of their human resources. Nowadays, Autonomous Mobile Robots (AMR) are state of the art technologies for such applications due to their great adaptability in dynamic environments, replacing more traditional solutions such as Automated Guided Vehicles (AGV), which are quite limited in terms of flexibility and require expensive facility updates for their installation. The application of Artificial Intelligence (AI) to increase AMRs capabilities has been contributing for the development of more sophisticated and efficient robots. Nevertheless, multi-robot coordination and cooperation for solving complex tasks is still a hot research line with increasing interest. This work proposes a Multi-Agent System for coordinating multiple TIAGo robots in tasks related to the manufacturing ecosystem such as the transportation and dispatching of raw materials, finished products and tools. Furthermore, the system is showcased in a realistic simulation using both Gazebo and Robot Operating System (ROS).
\end{abstract}

\section{\MakeUppercase{Introduction}}

Currently, the automation of internal logistics is a problem faced by both large and small corporations. By automating repetitive tasks such as material transportation and storing, companies are able to apply their human resources more effectively focusing on higher-value activities. 

Not so long ago, the only option to automate internal transportation tasks were Automated Guided Vehicles (AGV), first introduced in \citep{muller1983automated}, which are limited to pre-programmed software instructions and can only navigate through fixed paths. Their installation is also rather complex and typically requires extensive facility updates that can disrupt a company's activities, which is non-optimal. Recently, Autonomous Mobile Robots (AMR) have taken the place of AGV in many intralogistic operations due to their advanced hardware and control software which allows the automation of several operations in dynamic environments. Additionally, these are able to negotiate independently with other resources (machines, people, \textit{etc.}) contributing to a decentralized decision-making process in their operation \citep{FRAGAPANE2021405}.

The most recent advances in artificial intelligence are pushing the state of the art on intelligent robots further by improving capabilities such as path planning, dynamical control and trajectory tracking making them apt to be deployed in uncertain environments \citep{selfadapt}. As example, object detection algorithms such as deep Convolution Neural Networks (CNN) can be used to recognise specific types of objects and materials that should or should not be transported.

The ability for a robot to understand both its state as well as its surrounding environment and states of other robots can be crucial to make coherent decisions for completing complex tasks which require coordination and awareness of several resources \citep{MouadASKM11}. This situation can be expanded in terms of lone operator protection, since, for example, in the industrial context, AMRs can help to detect and early report incidents by being aware of their surroundings, hence, contributing to a better protection of the company's workers.

On the other hand, Multi-Agent Systems (MAS), a sub-field of Distributed Artificial Intelligence (DAI), is a well-known technology that can be applied to multi-robot coordination in the context of logistic and inventory-related tasks automation. A MAS is composed of several autonomous entities, agents, which are able to perceive their surrounding environment and work along side each other to interact with it \citep{8352646}.

In this work, we propose a MAS to coordinate the operation of multiple AMRs to automate several tasks regarding logistics and inventory activities in the manufacturing ecosystem such as moving and dispatching raw materials, finished products and tools. For our AMR model we used TIAGo \citep{Pags2016TIAGoTM}, a modular robot developed by PAL Robotics which provides perception, navigation, manipulation and Human-Robot Interaction (HRI) skills. The proposed system was tested in a realistic simulated environment built using the Gazebo \citep{gazebo} simulator.

This document is organized into multiple sections that can be detailed as follows. Section 2 describes artificial intelligence implementations in the context of AMRs, current AMR model developments and simulation environments. Section 3 explains the proposed MAS for AMR coordination in the manufacturing ecosystem. Section 4 presents the obtained results and their discussion. Section 5 provides the summary of our research main conclusions and appoints further research.

\section{\MakeUppercase{Related Work}}
\label{relatedwork}

The advent of automation is regarded as the future in manufacturing, with several works analysing and discussing this same topic \citep{FROHM2006453, human+}, citing Artificial Intelligence (AI) as one of its cornerstones.

Currently, the focus is on AMRs and the reduction of human interaction for their training/setup. These robots are expected to carry out a set of actions or tasks, while deployed in varied and not fully controlled environments. As such, they need to be able to detect and navigate around any unexpected obstacle. In \citep{FLOREANO19981461}, D. Floreano \textit{et al.} uses a combination of neural networks and evolutionary algorithms to investigate the possible use of biological behaviours in adaptive environment automation. The next step is an integration of the AMRs with a centralized control center, allowing greater collaboration between the different agents. In \citep{MAS}, T. Laengle \textit{et al.} describes the implementation of such a control center based on MAS, outlining its advantages of increased collaboration and scalability. Although MAS are receiving more and more attention due to their huge potential, they are not without downsides as discussed in \citep{8352646}, namely coordination between agents, security, and task allocation.

The difficulty and high requirements of robotics research and development creates a need for robust simulation frameworks. The differences between several available ones is already well documented in the literature \citep{article}, with the caveat that each of those tools were created with a specific purpose in mind. On the other hand, Robot Operating System (ROS) \citep{ros} is open source and endeavors for code re-use and high compatibility.
These tools facilitate the setup of a testing environment, allowing for fast prototyping of several different models of robots.

ROS relies on a network of code repositories, where different manufacturers can develop and release their own robot's models and software. This way a user can access and experiment with the most varied simulations and models, such as Boston Dynamics' famous Spot \citep{spot}.
Another interesting robot's models, also available for ROS, are namely PAL Robotics' TIAGo \citep{tiago1, tiago2}. TIAGo is a highly customizable robot, with great support for AI. TIAGo is employable in a myriad of task ranging from Factory of the Future applications along Smart Cities and Ambient Assisted Living as well as being a valuable asset for AMR research. Furthermore, by deploying multiple TIAGo robots one can achieve a higher level of cooperation to solve complex tasks.

\subsection{Simulation Environment}

ROS provides integration with a 3D modeling simulator, Gazebo \citep{gazebo}, which utilizes powerful physics engines and realistic 3D-graphics rendering to allow robot and environment model creation as well as the development of realistic simulations. 

\begin{figure}[H]
\centering
\includegraphics[width=8.5 cm]{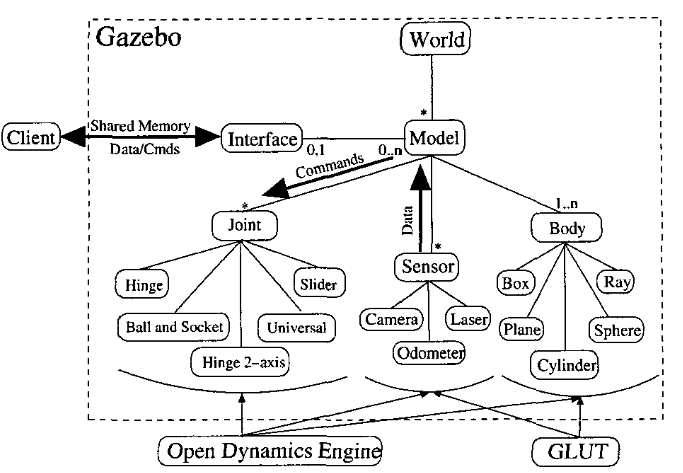}
\caption{Gazebo Components \citep{1389727}}
\label{fig:gaz}
\end{figure}

Driven by the increasing use of robotics in outdoor environments, Gazebo was first introduced in \citep{1389727} to allow a simple manner of modeling dynamic scenarios and provide realistic sensor feedback mechanisms. The general structure of Gazebo components is described in Figure \ref{fig:gaz}.

A Gazebo world is composed of multiple user created models that have the ability to interact with each other. All objects have numerous attributes such as mass, velocity and friction which allows them to behave realistically in the presence of an interaction such as pushing or pulling. On the other hand, the robots themselves are structures composed of rigid bodies which are connected by joints. To surfaces and joints, forces can be applied to create locomotion and interactions within the simulation. Additionally, sensibly every aspect of the simulation can be configured and controlled \citep{1389727}.

\subsection{Multi-Agent System Platforms}

In order to build a MAS capable of interfacing with ROS and controlling the behaviour of the robots in the simulation, several frameworks can be used, namely JAVA Agent DEvelopment Framework (JADE) \citep{bellifemine1999jade}, JADE eXtension (JADEX) \citep{jadex} and the more recent Smart Python Agent Development Environment (SPADE) \citep{9207929}, all of which are fully FIPA \citep{fipa} compliant. JADE has perhaps been the most popular platform for MAS development of the last two decades. It was fully implemented in the Java programming language, it is compliant with Foundation for Intelligent Physical Agents' (FIPA) specification and compatible across any operating system. JADEX is an addon for JADE that brings the BDI Archictecture \citep{BDI} philosophy and its goal-oriented  behaviour to JADE \citep{jadex}.

However, JADE's last version dates back to 2017, whereas the new-coming SPADE has been introduced to mitigate the open challenges of current agent platforms, such as providing a simple and effective communication channel, a higher resilience to unexpected growths in the number of messages and/or entities involved in the communication and to provide support independently of the device in which each agent is running. 

SPADE has an Agent Communication Channel (ACC) which is a component that manages all platform communications by establishing a connection with the XMPP server. This server, in turn, redirects all messages from the sender to the recipient without user intervention. This technology has several integrated security features such as user/password based authentication and SSL tunnel encryption. SPADE is also highly inter-operable since it allows the communication with any entity, such as virtual agents and human operators. It is also capable of establishing a connection with other MAS platforms such as JADE.

\section{\MakeUppercase{Proposed Method}}
\label{proposed}

To coordinate the operation of several AMRs in manufacturing-related activities we have developed a MAS using the Python programming language and SPADE framework \citep{9207929}. Two types of agents were considered, the \textit{Master} and the \textit{Robot} which communicate through a public XMPP server. The system was designed in a generic way and envisions one \textit{Master} agent and a variable number of \textit{Robot} agents. 

\begin{figure}[H]
\centering
\includegraphics[width=8 cm]{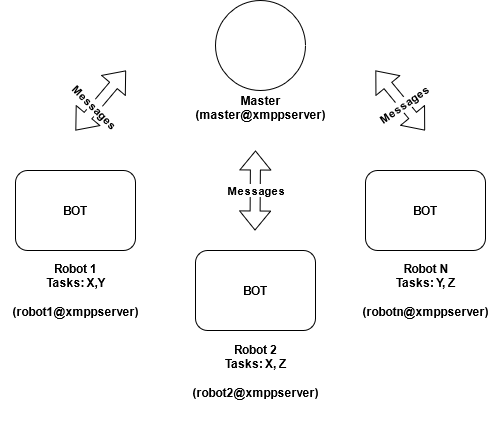}
\caption{Proposed Multi-Agent System.}
\label{fig:mas}
\end{figure}

The proposed MAS is represented in Figure \ref{fig:mas}.

The architecture of each of the mentioned Agent types is:

\begin{itemize}
    \item \textbf{Master:}
        \begin{itemize}
            \item \textbf{Static:} The \textit{Master} is a conceptual entity, hence it is not mobile.
            \item \textbf{Deliberative:} Possesses an internal symbolic, reasoning model and engages in planning and negotiation in order to achieve coordination with other agents.
            \item \textbf{Collaborative:} It cooperates with the other \textit{Robot} agents.
        \end{itemize}
    
    \item \textbf{Robot:}
        \begin{itemize}
            \item \textbf{Mobile:} It represents a physical robot which navigates through its environment.
            \item \textbf{Hybrid:} Agent is both deliberative and reactive as it will react to orders given by the \textit{Master} but will autonomously navigate through the environment.
            \item \textbf{Collaborative:} All \textit{Robot} agents cooperate with the \textit{Master} agent.
        \end{itemize}
\end{itemize}

Furthermore, each agent has its own account in the XMPP server contributing to a decentralized architecture, entirely independent from the physical location of where they are running. This way, the IP address of the machine does not need to be known by the rest of the agents in order for them to exchange different types of messages \citep{9207929}. The introduced agent types have distinct responsibilities. 

The \textit{Master} agent receives a list of products to be manufactured and orchestrates a variable number of robots in order to execute all the required tasks. On the other hand, the \textit{Robot} agent is able to perform a limited set of tasks and has the responsibility to communicate this information to the \textit{Master} when queried about it. Furthermore, he must perform the task assigned to him in a quick and safe manner. The task execution implies the autonomous navigation of the robot in the shop floor and collision detection mechanisms. When the robot faces an obstacle or a factory operator it must stop in order to prevent a collision and wait until its path is clear.

Each type of agent exchanges two different types of messages with distinct objectives:

\begin{itemize}
    \item \textbf{Identity}: The \textit{Master} agent receives a list of \textit{Robot} agents as argument in its constructor. These agents are the ones to be considered for task assignment. Nevertheless, they can be unavailable (\textit{e.g.} due to a maintenance intervention) or the tasks they are able to perform could have been changed. Therefore, the \textit{Master} starts by sending an \textbf{Identity Check} message to all known \textit{Robots} and waits for replies. The \textit{Robots} which receive the \textbf{Identity Check} reply with an \textbf{Identity} message, stating their identifier and which tasks they can execute.
    \item \textbf{Order}: When dispatching a given task to a \textit{Robot} the \textit{Master} sends an \textbf{Order} message containing the task identifier. On the other hand, when the \textit{Robot} finishes that task it replies with an \textbf{Order Notice} containing the total time it took to complete it.
\end{itemize}

Considering the presented architecture, two algorithms for the \textit{Master} agent's task processing behaviour were developed, one which attempts to dispatch each task sequentially and another who adopts balanced task queues for each available \textit{Robot}. The first iterates over every task and tries to dispatch it to a \textit{Robot} which is able to perform it. When there is no \textit{Robot} available, the \textit{Master} agent will hold until a capable \textit{Robot} finishes it's current task. The second algorithm starts by creating an empty queue for each available \textit{Robot}, then it first dispatches the tasks which can only be done by a single agent and finally, the remaining tasks are distributed in a balanced manner regarding the average time they take to complete. Many other algorithms could be implemented to solve this problem, so, the system was designed so they can be interchangeable without requiring significant changes in the code base. Hence, two design patterns were applied, \textit{Strategy} (behavioral) and \textit{Factory} (creational) \citep{gof}. 

\begin{figure}[H]
\centering
\includegraphics[width=6 cm]{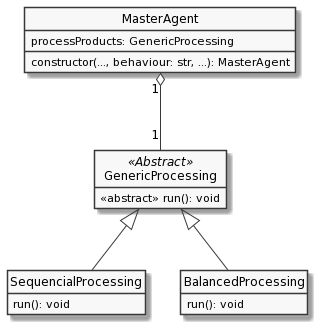}
\caption{Strategy design pattern applied to the task dispatching behaviour.}
\label{fig:strategy}
\end{figure}

Figure \ref{fig:strategy} presents a Unified Modelling Language (UML) class diagram which illustrates the application of the \textit{Strategy} pattern to the task dispatching behaviour. 

When an instance of the \textit{MasterAgent} class is created, one can specify a designation related to the intended task dispatching algorithm. The \textit{Factory} has the responsibility to create the correct concrete implementation according to the value of such designation.

\section{\MakeUppercase{Case Study}}
\label{case}

To demonstrate the potential of our solution, a simulation was created based on an industrial example, where the employment of robots to aid human workers is most common. The characterization of the considered environment is as follows:

\begin{itemize}
    \item \textbf{Dynamic}: As there are other entities acting in the same space, changes exists beyond the agent's control.
    \item \textbf{Inaccessible}: Due to its dynamic nature, there will be no way to obtain complete, accurate, up-to-date information about the environment's state.
    \item \textbf{Non-deterministic}: The agent will only affect a small part of the environment.
    \item \textbf{Continuous}: As the agent needs to adapt to the changing environment, there is an unlimited number of possible actions.
\end{itemize}

\begin{figure}[H]
\centering
\includegraphics[width=8 cm]{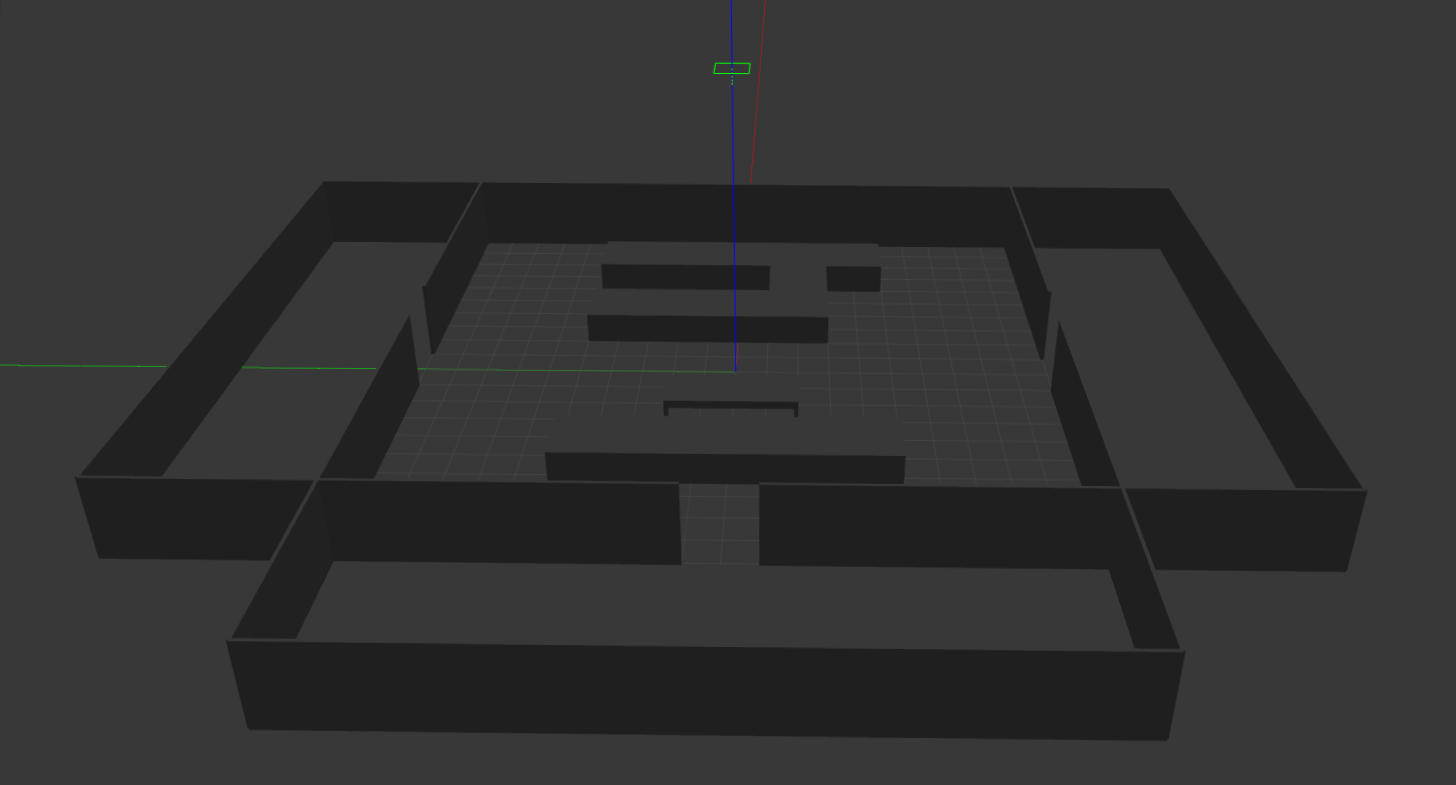}
\caption{Simulated factory floor.}
\label{fig:shop}
\end{figure}

Using ROS and Gazebo \citep{gazebo}, we constructed a 3D model of a factory floor. The main room is where the products are assembled and where the human operators will be working and circulating. To the bottom and sides of this main room are several depots from which the robots are tasked with retrieving tools and materials. Using Figure \ref{fig:shop} as reference, the left-side is where raw materials are located, bellow houses the tool rack and the right-side contains the conveyor belt where the finished product should be delivered.

Each product is comprised of a list of tasks. These tasks refer to courier runs to fetch the needed materials and tools required to assemble the product.
Our system is comprised of one \textit{Master} agent, to coordinate, and two \textit{Robot} agents, one for each of the real robots deployed. As the robots may only perform certain tasks, each agent contains a list for its robot such as: finished products and tools (\textbf{TIAGo1}) and raw materials and tools (\textbf{TIAGo2}).

For testing purposes \textbf{ProductX} will be used, comprised of the following lists of tasks:
    
\begin{itemize}
    \item Setup: preparation of the floor for product manufacturing, by delivering Material1, Material2, Tool1, Tool2.
    \item Cleanup: clear the floor, by storing ProductX, Tool1, Tool2.
\end{itemize}

The \textit{Master} agent uses two different algorithms to distribute the tasks: Sequential and Balanced.

Applying the sequential one, the agent will iterate through the list and attempt to assign each task, waiting when no robot is available. For our \textbf{ProductX} this means that the first task of "Material1" is assigned to the available robot: TIAGo2. As the next task is also of type material and the only remaining TIAGo is not capable of completing those types of tasks it will wait until TIAGo2 finishes its current task to assign "Material2".

Whereas the balanced algorithm solves this issue by generating a unique queue for each of the available robots, distributing the tasks to each queue taking into account robot-task compatibility and the overall time each one will take to complete. In our case, since only one of the TIAGos fetches materials, the queues are:

    \begin{itemize}
        \item \textbf{Queue-TIAGO1:} Tool1, Tool2.
        \item \textbf{Queue-TIAGO2:} Material1, Material2.
    \end{itemize}

The \textit{Robot} agent uses laser-scan sensor data to identify obstacles in its path and avoid collisions. It will wait a certain amount of time for the obstacle to clear before proceeding with its assigned task.

\begin{figure}[H]
\centering
\includegraphics[width=8 cm]{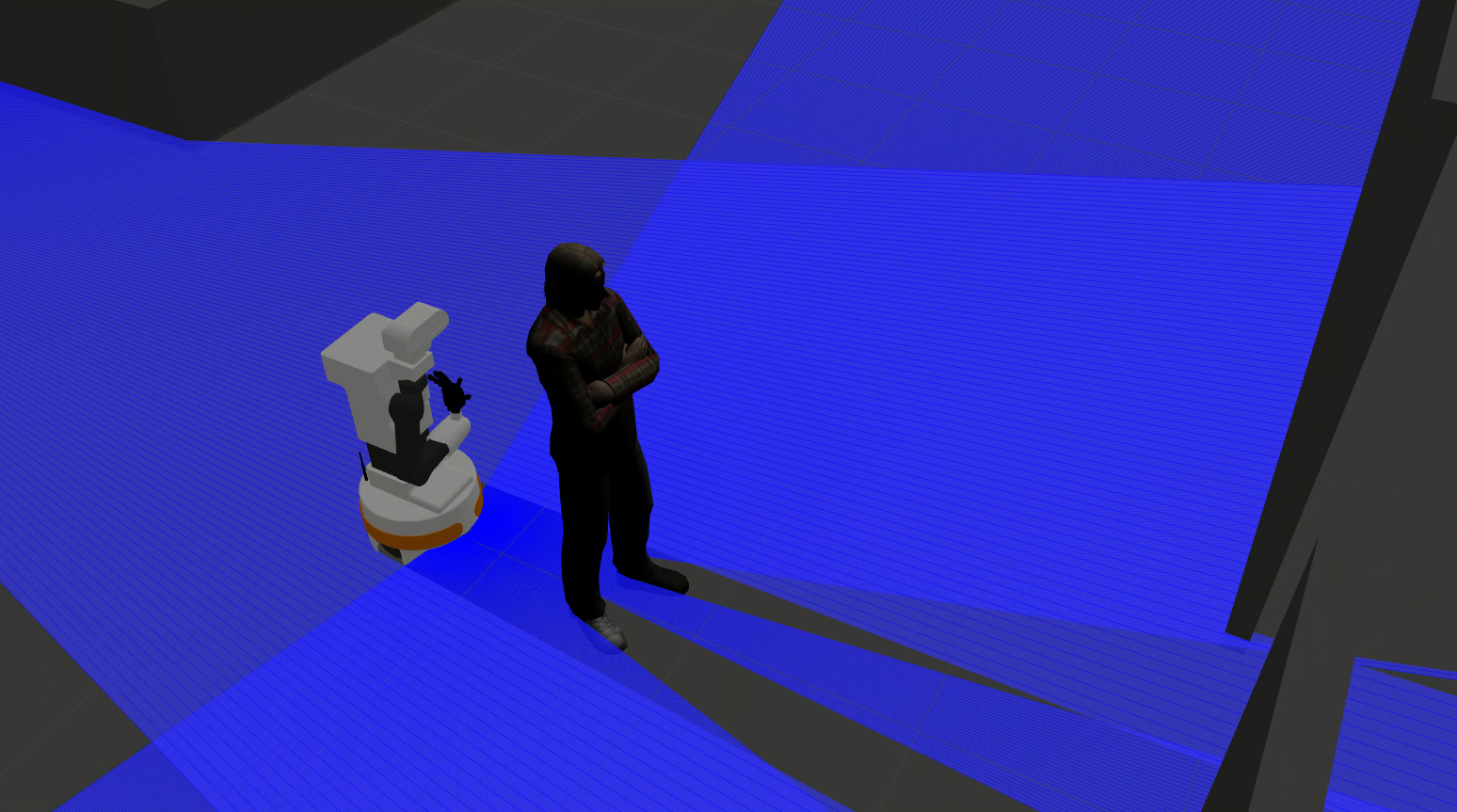}
\caption{Obstacle detection.}
\label{fig:stop}
\end{figure}

\section{\MakeUppercase{Conclusion}}
\label{conclusion}
Modern corporations face, on a daily basis, several challenges regarding the automation of logistics and inventory related activities. Nowadays, solutions based on AMRs are cost-efficient technologies for such applications exhibiting great adaptability in dynamic environments. This work proposes a scalable MAS for coordinating a variable number of AMRs in complex tasks related to product manufacturing. The proposed system provides a simple way to interchange between task dispatching algorithms and an extensible code base where one can easily add new implementations of such algorithms without requiring substantial changes.

Furthermore, the proposed system was showcased in a reliable case study which was setup using ROS and Gazebo. It regards, a shop floor with three rooms, for raw materials, tools and finished products.  In this environment, two TIAGo robots were deployed and controlled using our MAS and ROS to manufacture, in a simulated manner, several products according to two different algorithms. For each assigned task the robot had to move to the correct room and avoid any collision in its way. This mechanism was implemented using the laser scan of the robot's mobile base.

In the future, additional features regarding autonomous navigation and collision detection will be implemented and more complex case studies will also be considered. Additionally, human-robot interaction will be explored by applying computer vision algorithms to the images collected by the camera present in the TIAGo robot's head.

\section{\MakeUppercase{Acknowledgements}}

The present work was done and funded in the scope of European Union's Horizon 2020 research and innovation programme under project SeCoIIA (grant agreement No 871967). This work has also received funding from the following projects: UIDB/00760/2020 and CyberFactory\#1 (Refª: NORTE-01-0247-FEDER-40124). This output reflects the views only of the authors, and the European Union cannot be held responsible for any use which may be made of the information contained therein.

\bibliography{refs}

\end{document}